\pdfoutput=1
\documentclass[12pt]{article}
\usepackage[utf8]{inputenc}
\usepackage[T1]{fontenc}
\usepackage[sorting=none]{biblatex} 
\bibliography{main}
\usepackage[utf8]{inputenc}
\usepackage{longtable}
\usepackage{blindtext,alltt}
\usepackage{indentfirst}
\usepackage{hyperref}
\usepackage{graphicx}
\usepackage{amsmath}
\usepackage{tabulary}
\usepackage{booktabs}
\usepackage{algorithm}
\usepackage{algpseudocode}
\usepackage[skip=2pt]{caption}
\usepackage{amssymb}
\usepackage{relsize}
\usepackage{authblk}
\usepackage{url}
\usepackage[small]{titlesec}
\usepackage[a4paper, total={6.7in, 10.1in}]{geometry}
\usepackage{mathtools}

\DeclareMathAlphabet{\mathbfit}{OML}{cmm}{b}{it}

\renewcommand{\thefootnote}{\fnsymbol{footnote}}
\newcommand\blfootnote[1]{%
  \begingroup
  \renewcommand\thefootnote{}\footnote{#1}%
  \addtocounter{footnote}{-1}%
  \endgroup
}

\providecommand{\keywords}[1]
{
  \small	
  \textbf{\textit{Keywords---}} #1
}
\date{}
\begin{document}

\title{
\Huge{Bottom-up Iterative Anomalous Diffusion Detector (BI-ADD)}\\
}

\author[1\thanks{Corresponding author}]{Junwoo Park}
\author[1]{Nataliya Sokolovska}
\author[2]{Cl\'ement Cabriel}
\author[2]{Ignacio Izeddin}
\author[1]{Judith Min\'e-Hattab}
\affil[1]{\small{Sorbonne Universit{\'e}, CNRS, Laboratoire de Biologie Computationnelle et Quantitative, LCQB, F-75005 Paris, France}}
\affil[2]{\small{Institut Langevin, ESPCI Paris,
Universit{\'e} PSL, CNRS, Paris, France}}

\maketitle

\begin{abstract}
   In recent years, the segmentation of short molecular trajectories with varying diffusive properties has drawn particular attention of researchers, since it allows studying the dynamics of a particle. In the past decade, machine learning methods have shown highly promising results, also in changepoint detection and segmentation tasks. Here, we introduce a novel iterative method to identify the changepoints in a molecular trajectory, i.e., frames, where the diffusive behavior of a particle changes. A trajectory in our case follows a fractional Brownian motion and we estimate the diffusive properties of the trajectories. The proposed BI-ADD\blfootnote{This is the version of the article before peer review or editing, as submitted by an author to \href{https://iopscience.iop.org/journal/2515-7647}{JPhys Photonics}. IOP Publishing Ltd is not responsible for any errors or omissions in this version of the manuscript or any version derived from it.} combines unsupervised and supervised learning methods to detect the changepoints. Our approach can be used for the analysis of molecular trajectories at the individual level and also be extended to multiple particle tracking, which is an important challenge in fundamental biology. We validated BI-ADD in various scenarios within the framework of the AnDi2 Challenge 2024 dedicated to single particle tracking. Our method is implemented in Python and is publicly available for research purposes.
\end{abstract}

\keywords{Single Particle Tracking (SPT), anomalous diffusion, fractional Brownian motion, AnDi2 challenge 2024, trajectory segmentation, neural network}

\section{Introduction}
Finding a pattern in a continuous random process to predict the future from limited observations has been extensively studied to predict stock index, consumer's patterns in economics, trajectories of molecules in physics and fractals in mathematics~\cite{hurst_math, husrt_finance, hurst_finance2}.
In 1951, the British hydrologist Harold Edwin Hurst introduced the Hurst exponent $H$ to determine the dam size of the Nile river depending on the periodical rain observed in Egypt~\cite{Husrt1951}. Since then, various methods have been developed to estimate $H$, the long-range dependence in a random process, from classical methods such as rescaled range (R/S) analysis~\cite{Husrt1951} to machine learning based methods~\cite{anomalous_Gentili,anomalous_Munoz-Gil,anomalous_quiblier,anomalous_Verdier}, with the goal to estimate $H$ as accurate as possible.

In biology, 
with the recent development of single molecule microscopy~\cite{betzig2006, betzig2008}, it became possible to rapidly image the dynamics of thousands of proteins in living cells. By quantifying the dynamics of protein populations, we can observe protein interactions or explore the activity of novel pharmaceuticals with previously unmatched resolution.
The molecular diffusion can be described by the anomalous diffusion coefficient $\alpha$ (where $\alpha=2H$) and the generalized diffusion coefficient $K$; $\alpha$ and $K$ represent the dependence of molecular diffusion over time and the magnitude of diffusivity respectively. More precisely, the anomalous exponent $\alpha$ represents the degree of recurrence of DNA exploration, that is, the number of times a DNA locus re-iteratively scans neighboring regions before reaching a distant position. When $\alpha$ is low, the locus explores recurrently the same environment for a long time reaching the same targets, while a high $\alpha$ indicates that the locus is able to explore new environments often. However, the prediction of $\alpha$ still remains challenging mainly due to the short length of molecular trajectories coming from experimental limitations.

The low number of observations is a major hurdle which limits the improvement of $\alpha$ estimation and further analysis. The physical limitations of empirical methods in optics and noise in low-resolution images of molecules also increase uncertainty in statistical results. Thus, the accurate estimation of $\alpha$ from short sequences is absolutely needed for the prediction of molecular dynamics and its change in a heterogeneous environment.

In our contribution, we introduce a method to estimate the $\alpha$ and $K$ with neural networks from a individual short molecular trajectory following fBm (fractional Brownian motion)~\cite{Mandelbrot1968}~\cite{fbm_Falconer} which is a generalization of Brownian motion having auto-covariance function given as:
\begin{equation}\label{eq:12}
	\mathbfit{E}[B_H(t)B_H(s)] = \frac{1}{2}(|t|^{\alpha} + |s|^{\alpha} - |t - s|^{\alpha}), 
\end{equation}
where $t \in \mathbb{R}$ and $\alpha \in (0, 2)$. The accurate estimation of $\alpha$ and $K$ is crucial to determine the \textit{\textbf{changepoint}} in a trajectory, which is the moment where molecular dynamics changes are induced by changes of $K$ and/or $\alpha$. These changes can reflect the activity of a molecule, such as the binding to a substrate, a sudden change of molecular crowding when entering or exiting a macro-domain. 

The classical method to analyze molecular properties at ensemble-level in homogeneous systems is Mean Squared Displacement (MSD)~\cite{LangevinMSD}: 
\begin{equation}\label{eq:11}
	\text{MSD}(t) = 2nKt^{\alpha}, 
\end{equation}
where $n$ is the dimension.
In biology, the MSD represents the amount of space a locus has explored in the nucleus, cytoplasm and membrane which can reveal the nature of molecular motion at ensemble-level. When molecules freely diffuse ($\alpha$=1), its MSD curve is linear in time and its motion is called Brownian. However, in living cells, the molecular motion is often slower than Brownian diffusion and is called subdiffusive ($0<\alpha<1$)~\cite{barkai2012strange} and the future trajectory is negatively correlated to the past trajectory. Several types of subdiffusive motion have been observed in biological tasks~\cite{Mine-Hattab2017-in}. When a chromosomal locus is confined inside a sub-volume of the nucleus, the motion is called confined subdiffusion and the MSD exhibits a plateau~\cite{marshall1997interphase}. When a force or structure that restricts the motion is not a simple confinement but is modulated in time and space with scaling properties, the motion is called anomalous subdiffusion~\cite{barkai2012strange,metzler2014anomalous}. The superdiffusion ($1<\alpha<2$) usually indicates that the molecules are able to explore new environments and the future trajectory is positively correlated to the past trajectory. However, MSD is not a proper method in a heterogeneous environment, since it averages over the ensemble of trajectories without considering the change of dynamics, and does not fit well in general. 

To analyze the molecular trajectory at individual level and to approximate the changepoints, we propose a novel method to classify molecular trajectories by dividing each trajectory into multiple sub-trajectories. We introduce  Bottom-up Iterative Anomalous Diffusion Detector (BI-ADD) which identifies changepoints of molecular trajectories at individual level when molecules diffuse under fBm. BI-ADD integrates both classical and deep learning approaches for the analysis of molecular trajectories. The predicted changepoints with BI-ADD can indicate the significant changes of $\alpha$ and/or $K$ along the trajectory.

The AnDi Challenge~\cite{andi2_dataset,andi2_manuscript,andi1_paper} aims to analyze biological phenomena with inspections of molecular trajectories in cells. The molecular trajectories can be studied collectively in a homogeneous system or individually in a heterogeneous system if the diffusive properties change over time and space. The 2nd Anomalous Diffusion Challenge (AnDi2 Challenge), which is a successive challenge of 1st challenge~\cite{andi1_paper}, has been held in 2024 to answer the biological questions in terms of molecular trajectory and we participated in the challenge as SU-FIONA team. The performance of BI-ADD compared to other methods participated in the challenge is available in~\cite{andi2_manuscript,andi_challenge_result}; we were ranked 1st for the video trajectory task and 6th for the trajectory task (\url{http://andi-challenge.org/challenge-2024/#andi2leaderboard}). Since the BI-ADD detects the changepoints and the diffusive properties from the molecular trajectory only, we applied FreeTrace~\cite{FreeTrace} to infer the trajectory of molecules from microscopy videos of AnDi2 challenge. The simulated fBm trajectories to train the deep learning models and also to evaluate the obtained results were provided by the AnDi2 challenge~\cite{andi2_dataset}.

\section{Methods}

\begin{table}[htbp]\caption{Notations}
    \begin{center}
        \begin{tabular}{r p{14cm}  p{1cm}}
            \toprule
            Symbol & Description & Units\\
            \midrule
            $\mathbf{X}$ & x coordinate sequence of trajectory &pixel\\
            $\mathbf{Y}$ & y coordinate sequence of trajectory &pixel\\
            $\alpha$ & anomalous diffusion exponent &\\
            $K$ & generalized diffusion coefficient &$\frac{pixel^2}{frames^{\alpha}}$\\
            $T$ & length of trajectory &frames\\
            $T_{sub}$ & length of sub-trajectory &frames\\
            $cp$  & changepoint  & frame\\
            $\mathbf{R}$ & radial displacement sequence &pixel\\
            $\lambda$ & changepoint threshold for $\mathbf{S}$& \\
            $N$ & number of changepoints in a trajectory &\\
            $k$ & number of different states (clusters) in a sample&\\
            $\mathbf{X}^t_{w_i}$ & $t^{th}$ element inside a sliding window of size $w$ at $i^{th}$ position of $\mathbf{X}$ &\\
            $\sigma_{\mathbf{X}^{w}_{i}}$ & standard deviation of sliding window of size $w$ at $i^{th}$ position of $\mathbf{X}$ &\\
            $\mathbf{V}^{w}$ & converted signal, generated with sliding window of size $w$ &\\
            $\mathbf{S}$ & signal averaging $\mathbf{V}^{w}$ with $w_{set}$ &\\
            $w_{set}$ & set of sliding window sizes &\\
            $l$ & extension length for each first and last frames&\\
            $N_{sub}$ & number of sub-trajectories divided for the regression of $\alpha$ &\\
            $\mathbb{T}$ & set of input trajectory lengths for the Model$_{\alpha}$ &\\
            \bottomrule
        \end{tabular}
    \end{center}
    \label{tab:notation_table}
\end{table}

\subsection{Workflow overview}
The main objective of BI-ADD is two-fold: 
\begin{enumerate}
 \item Changepoints detection along the trajectory.
 \item Estimation of the molecular diffusive properties $\alpha$ and $K$ of sub-trajectories divided by the changepoints in a heterogeneous molecular system. 
 \end{enumerate}
 
\begin{figure}[!htb]
    \minipage{0.99\textwidth}%
    \centering
    \includegraphics[scale=0.95]{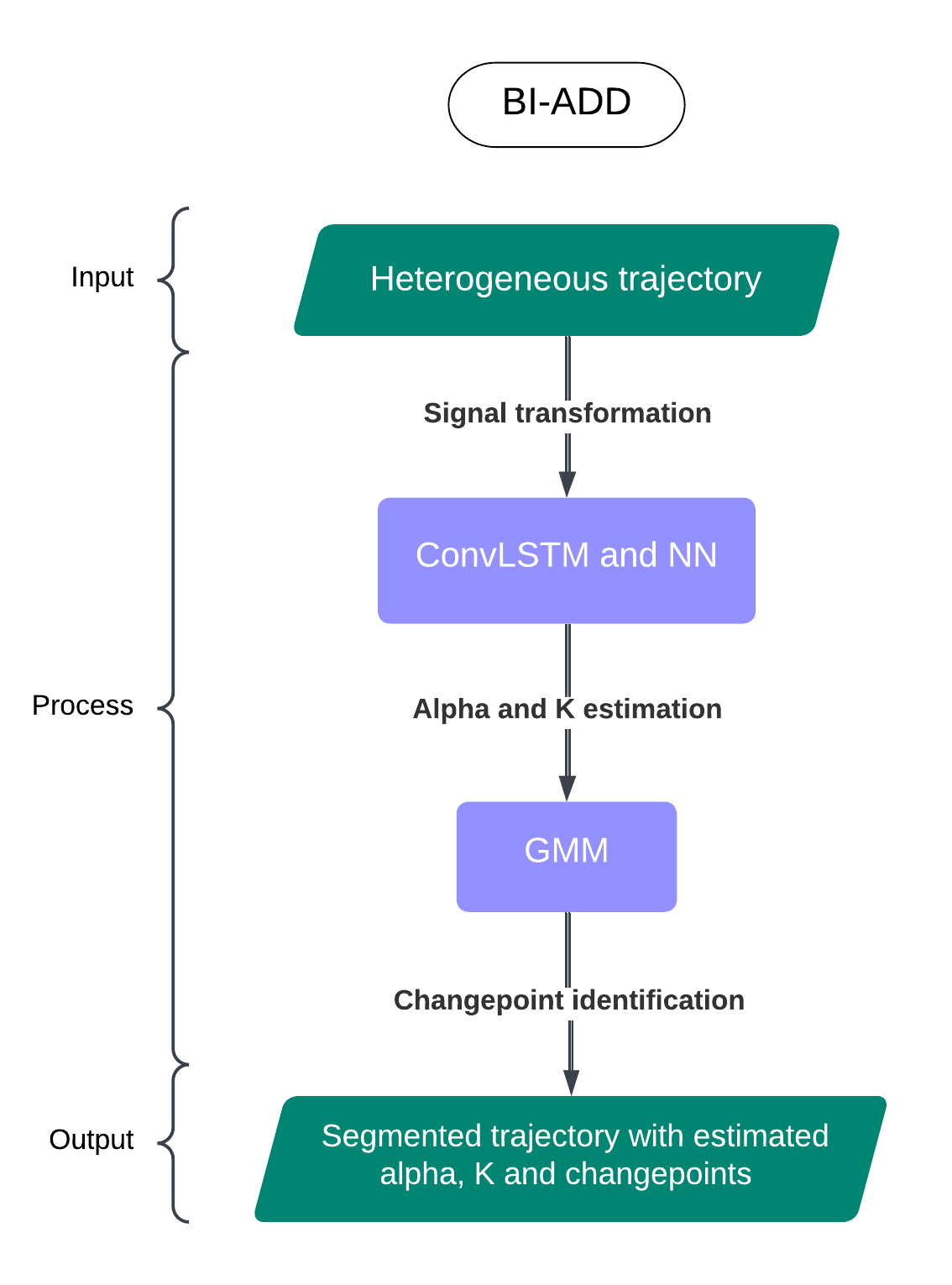}
    \endminipage
\caption{Simplified workflow of BI-ADD. The input is a sample of heterogeneous trajectories changing their diffusivity over time. BI-ADD transforms a trajectory into a signal with multiple sizes of sliding windows. The transformed signal suggests potential changepoints for each individual trajectory where the diffusivity changes significantly over time. The estimation of $\alpha$ and $K$ for sub-trajectories divided by changepoints and subsequent clustering filter the false positive changepoints. In the end, BI-ADD regenerates segmented trajectories from a given sample with their estimated diffusive properties. More detailed workflow is available in Figure S2.}
\label{fig:flowcharts}
\end{figure}

 First, we generate $N$ candidate changepoints by converting coordinates of a trajectory into a signal in the Euclidean space. Second, $\hat{\alpha}$ and $\hat{K}$ of $N+1$ sub-trajectories which are generated by dividing the trajectory with the potential changepoints are estimated. The diffusive properties are estimated with the regressions, performed with ConvLSTM (Convolutional Long short-term memory)~\cite{lstm,convneuralnetwork,convlstm} and a neural network (NN)~\cite{deeplearning} for $\hat{\alpha}$ and $\hat{K}$ respectively. Third, BI-ADD clusters $\hat{\alpha}$ and $\hat{K}$ from the entire sample to determine the total number of different states of the trajectories with the Gaussian Mixture Model (GMM)~\cite{scikit-learn}. Finally, BI-ADD re-visits each individual sub-trajectory iteratively to determine whether two consecutive sub-trajectories should be merged. This decision is taken based on the  clustering results and the proximity of two sub-trajectories given a fixed threshold. If BI-ADD merges two consecutive sub-trajectories, it re-estimates the $\hat{\alpha}$ and $\hat{K}$ of the merged new sub-trajectory and repeats the procedure until there are no more unexplored sub-trajectories. The workflow is shown on Figure~\ref{fig:flowcharts}.

\begin{figure}[!htb]
    \minipage{0.99\textwidth}%
    \centering
    \includegraphics[scale=0.31]{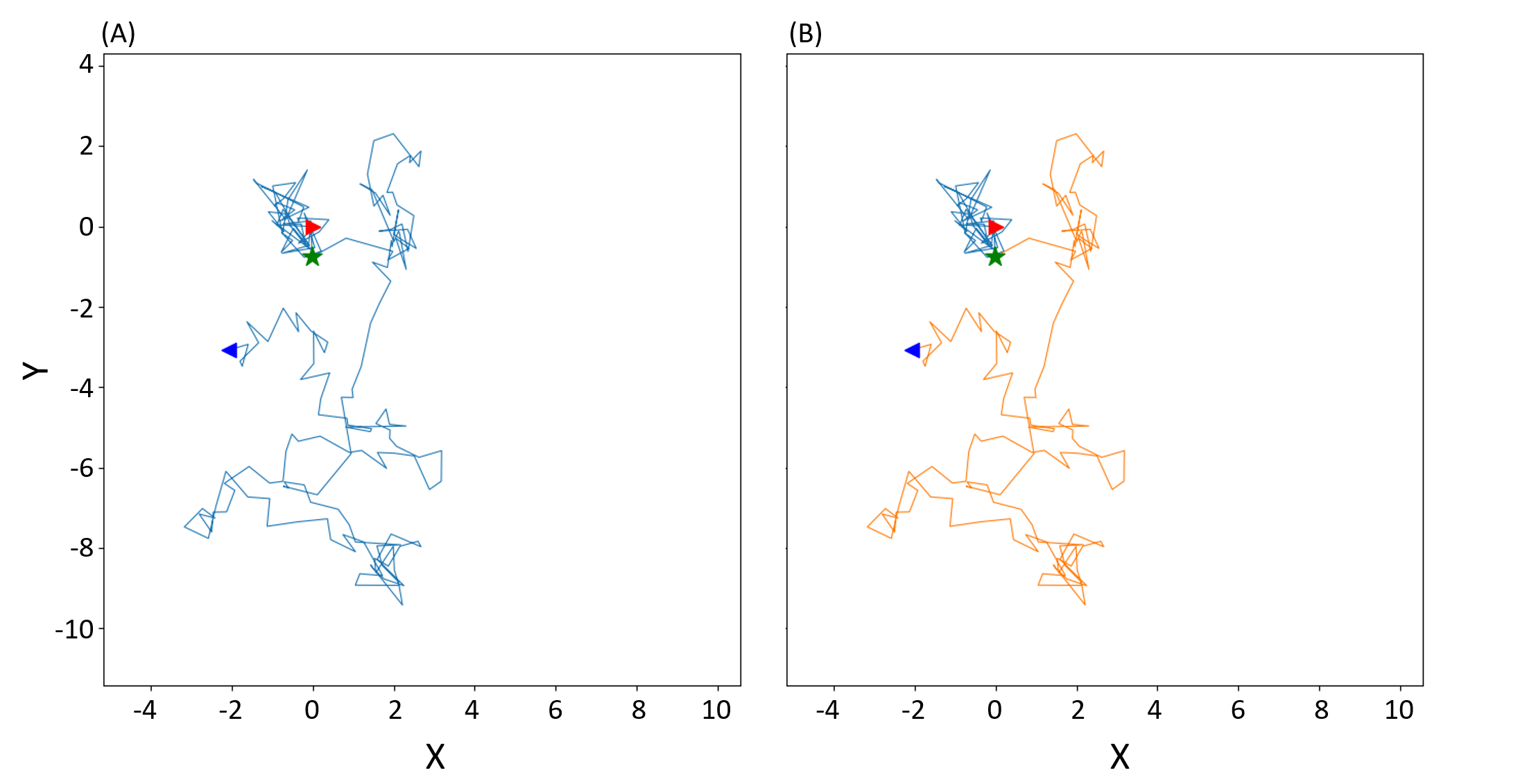}
    \endminipage
\caption{An example of two-dimensional fBm trajectory, where $\alpha$ changes from 0.2 to 1.0 while $K$ remains constant. The red and blue triangles indicate the start and end points of the trajectory respectively. The green star indicates the ground truth changepoint. The length of the simulated fBm trajectory is 200 and the $\alpha$ changes at the 49th frame. (A) Original trajectory; (B) Two sub-trajectories in orange and blue divided by the identified changepoint.}
\label{fig:fbm_trajectory}
\end{figure}

\subsection{Detection of potential changepoints}

\textit{Data}. A 2-dimensional trajectory (Figure~\ref{fig:fbm_trajectory}) of length $T$ is a sequence of coordinates $\{(x_1,y_1), \dots, (x_T, y_T)\}$. 

\textit{Assumption~1.} Given a trajectory, a changepoint divides the trajectory into two sub-trajectories. In this case, the sub-trajectories will belong to two different clusters, where the clusters are characterised by the diffusive  properties $\alpha$ and $K$.

\textit{Assumption~2.} The clusters described by the diffusion properties can be seen as the states of a Markov process, where the transition probability matrix describes the probability of molecular dynamics change in a trajectory.

\smallskip

Since fBm is a continuous Gaussian process, we built a distribution of $\hat{K}$ and $\hat{\alpha}$ to estimate the total number of clusters (states) $k=|\{(\hat{K}_n, \hat{\alpha}_n)|n\in\{1, 2, ..., k\}\}|$. The estimated $\hat{K}$ and $\hat{\alpha}$ sets in a given sample filter out false positive changepoints coming from noise; small fluctuations of $\hat{\alpha}$ and $\hat{K}$ are not frequently observed in the sample. To identify the clusters, we find potential changepoints candidates which divide a trajectory into multiple sub-trajectories including false positives. The advantage of this approach is the low number of false negatives which leads to a minimal regression error.

First, we extend a trajectory with a length of $2l$ by reflecting the extremities where the first and last coordinates are used as pivot points in the 2-dimensional Euclidean space. This extension of a trajectory is needed to approximate the changepoints, to estimate $\hat{\alpha}$ and $\hat{K}$ of sub-trajectories which are near the extremities of the original trajectory. Next, we convert the extended 2-dimensional trajectory into a signal using multiple sliding windows with the set of window sizes $w_{set}=\{20, 22, ..., 40\}$, i.e. the different numbers of frames considered, as follows:

\begin{equation}\label{eq:3_3}
	\displaystyle \mathbf{X}^{w}_{i,l} = \bigg\{ \mathbf{X}_{w_i}^k  - \mathbf{X}_{w_i}^0 \mid k \in [0, w/2] \bigg\}
\end{equation}
\begin{equation}\label{eq:3}
	\displaystyle A^{w}_{i,l} = \sum \left| \mathbf{X}^{w}_{i,l}\right|
\end{equation}
where $\mathbf{X}^k_{w_i}$ corresponds to the \textit{k-th} element of extended \textbf{X}, sliced with a sliding window of size \textit{w} at \textit{i-th} position of extended sequence. The subscripts $l$ and $r$ of equations~(\ref{eq:3_3})~(\ref{eq:3})~(\ref{eq:5}) stand for the range of $k \in [0, w/2)$ and $k \in [w/2, w)$ which correspond to the first and second half inside the sliding window respectively. The idea behind equation (\ref{eq:3_3}) is to identify the maximal distance a molecule can move in a given time.  
$A^{w}_{i,l}$ corresponds to the last value in an absolute cumulative sum inside a first half of the sliding window of size $w$ in $X$ space. $B^{w}_{i,l}$ is computed similarly to $A$ for $Y^{w}_{i,l}$ space. 

{\footnotesize
\begin{equation}\label{eq:5}
	\displaystyle v_i^w =  \bigg|\ \frac{A^{w}_{i,l} - A^{w}_{i,r}}{max(A^{w}_{i,l}, A^{w}_{i,r})} + 
\frac{B^{w}_{i,l} - B^{w}_{i,r}}{max(B^{w}_{i,l}, B^{w}_{i,r})}\ \bigg|  +  \delta
\end{equation}
where $\delta = \lvert\ \sigma_{\mathbf{X}^{w}_{i,l}} - \sigma_{\mathbf{X}^{w}_{i,r}}\ \rvert + \lvert\ \sigma_{\mathbf{Y}^{w}_{i,l}} -  \sigma_{\mathbf{Y}^{w}_{i,r}}\ \rvert$, $\sigma$ corresponds to the standard deviation. 
} 

The main intuition behind this transformation is that the newly constructed signal reflects the relative $\alpha$ and $K$ differences of two sub-sequences of size $w/2$ inside the sliding window. Note that equation (\ref{eq:5}) can be extended to a higher-dimensional trajectory since the transformed signal is a linear combination of simple terms.

The $\mathbf{V}^w$ is a signal representing relative difference of sub-sequences inside sliding windows measured for every $i$ except extended range:
\begin{equation}\label{eq:2}
\mathbf{V}^w = \bigg\{v_i^w | i\in [\,l, l+T]\bigg\}
\end{equation}

We sum $\mathbf{V}^w$ over all possible window sizes to get $\mathbf{S}$ -- our signal of interest -- the signal integrating information of all considered sliding windows $w$ on multiple scales:
\begin{equation}\label{eq:1}
\mathbf{S} = \underset{w\in w_{set}}{\mathlarger{\mathlarger{\mathlarger{\Sigma}}}} \mathbf{V}^w,
\end{equation}
where $i, j, l \in \mathbb{N}$, $w \in 2\mathbb{N}$. Note that the additional points added on the extremities do not influence the result since they are excluded from equation (\ref{eq:2}).  

\begin{figure}[!htb]
    \minipage{0.99\textwidth}%
    \centering
    \includegraphics[scale=0.3]{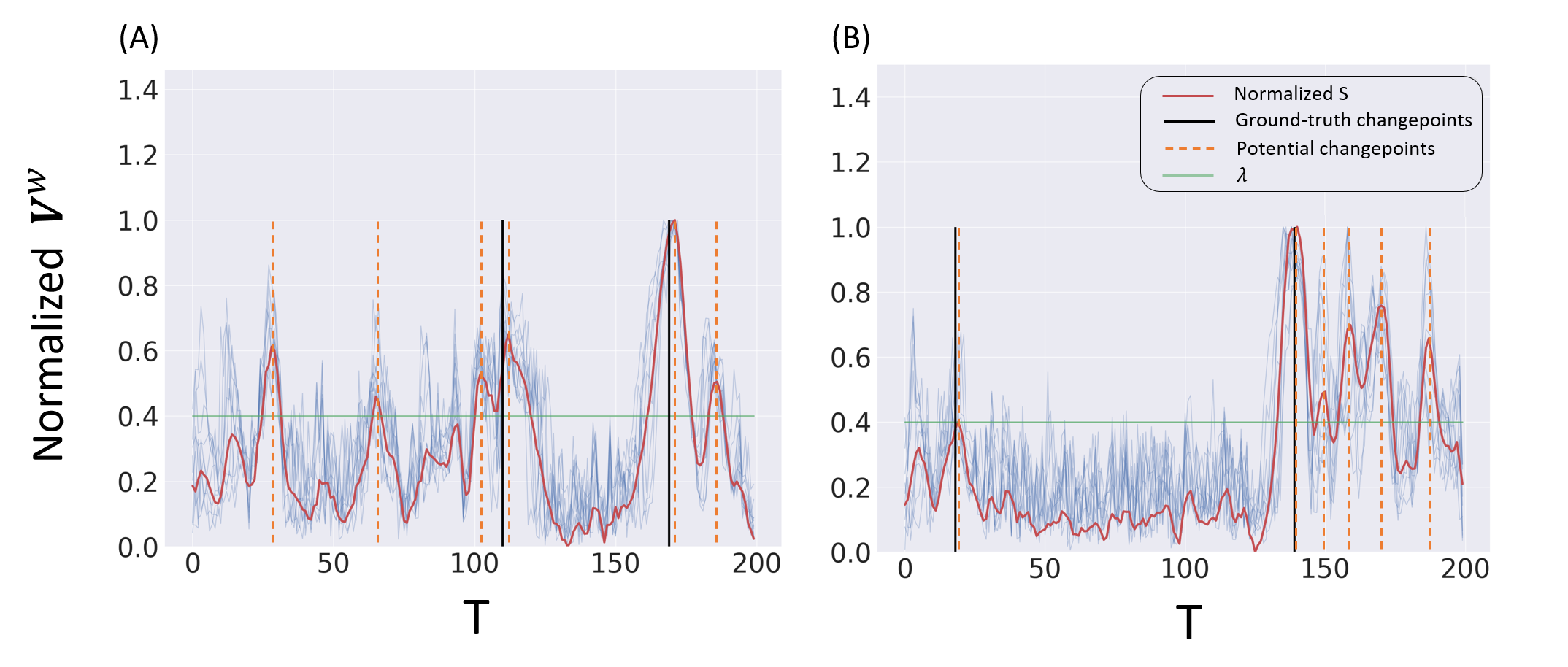}
    \endminipage
\caption{A normalized signal $\mathbf{S}$ computed with multiple sizes of sliding windows on two simulated trajectories. The black vertical lines show the ground-truth changepoints for each simulated trajectory. The orange dashed vertical lines show the potential changepoints above $\lambda$. The green horizontal line corresponds to the threshold value $\lambda=0.4$. The red signal is normalized $\mathbf{S}$. The transparent blue curves correspond to the $\mathbf{V}^w$ with various sizes of sliding window, where $w\in w_{set}$. The fBm trajectory of (A) is simulated with two states $(K_1,\alpha_1)=(0.01, 0.2)$ and $(K_2,\alpha_2)=(0.1, 1.0)$. (B) is simulated with $(K_1,\alpha_1)=(0.05, 0.5)$ and $(K_2,\alpha_2)=(0.1, 1.5)$. The transition probability between the states for both (A) and (B) is 0.01. The peaks of the signal $\mathbf{S}$ indicate the detected potential changepoints along the trajectory, where molecular dynamics changes.}
\label{fig:biadd_signals}
\end{figure}

The choice of the sliding window size depends on the transition probability between sub-trajectories -- these can be seen as Markov states -- in a heterogeneous system: a small size of a sliding window fits well with the high transition probability, large sizes of sliding windows reflect the case of low probability of the transition. An example of a normalized signal $\mathbf{S}$ is illustrated on Figure~\ref{fig:biadd_signals}. We can observe that a threshold $\lambda$ divides a trajectory into multiple sub-trajectories by finding the local maxima lying on the normalized signal $\mathbf{S}$. The local maxima above $\lambda$ can be considered as potential changepoints where the relative difference is highest between two sub-sequences. We can approximate potential changepoints including false positives from the normalized $\mathbf{S}$ and can estimate $\hat{\alpha}$ and $\hat{K}$ for each individual sub-trajectory.

\subsection{Estimation of $\hat{\alpha}$ and $\hat{K}$ for each sub-trajectory}
\begin{figure}[!htb]
    \minipage{0.99\textwidth}%
    \centering
    \includegraphics[scale=0.3]{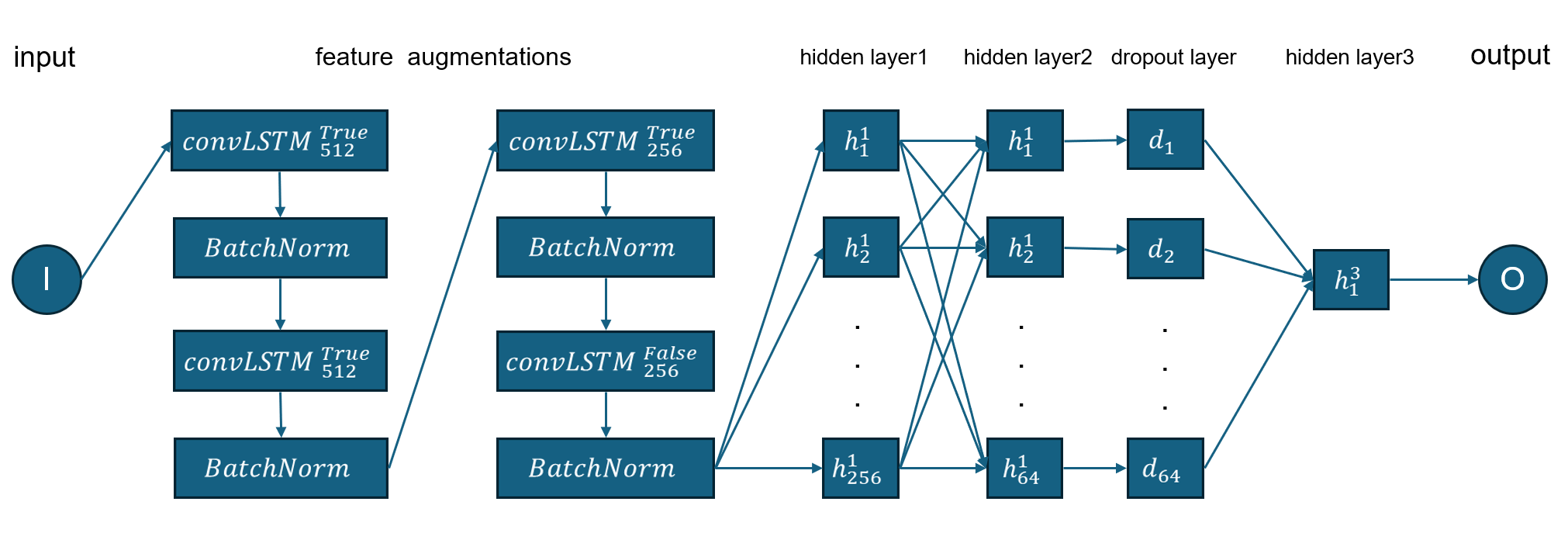}
    \endminipage
\caption{Model$_{\alpha}$ architecture: estimation of $\hat{\alpha}$. The conv$LSTM^{True}_{512}$ stands for returning a full sequence with 512 convolutional filters, the last output otherwise; $h^{k}_{n}$ represents the $n^{th}$ neuron in $k_{th}$ hidden layer. This model takes 3 input features (equations \ref{eq:6},~\ref{eq:7},~\ref{eq:8}). The convLSTM layers perform the feature augmentation. The total number of parameters in the model is 17M.}
\label{fig:alpha_reg_arch}
\end{figure}

\begin{figure}[!htb]
    \minipage{0.99\textwidth}%
    \centering
    \includegraphics[scale=0.3]{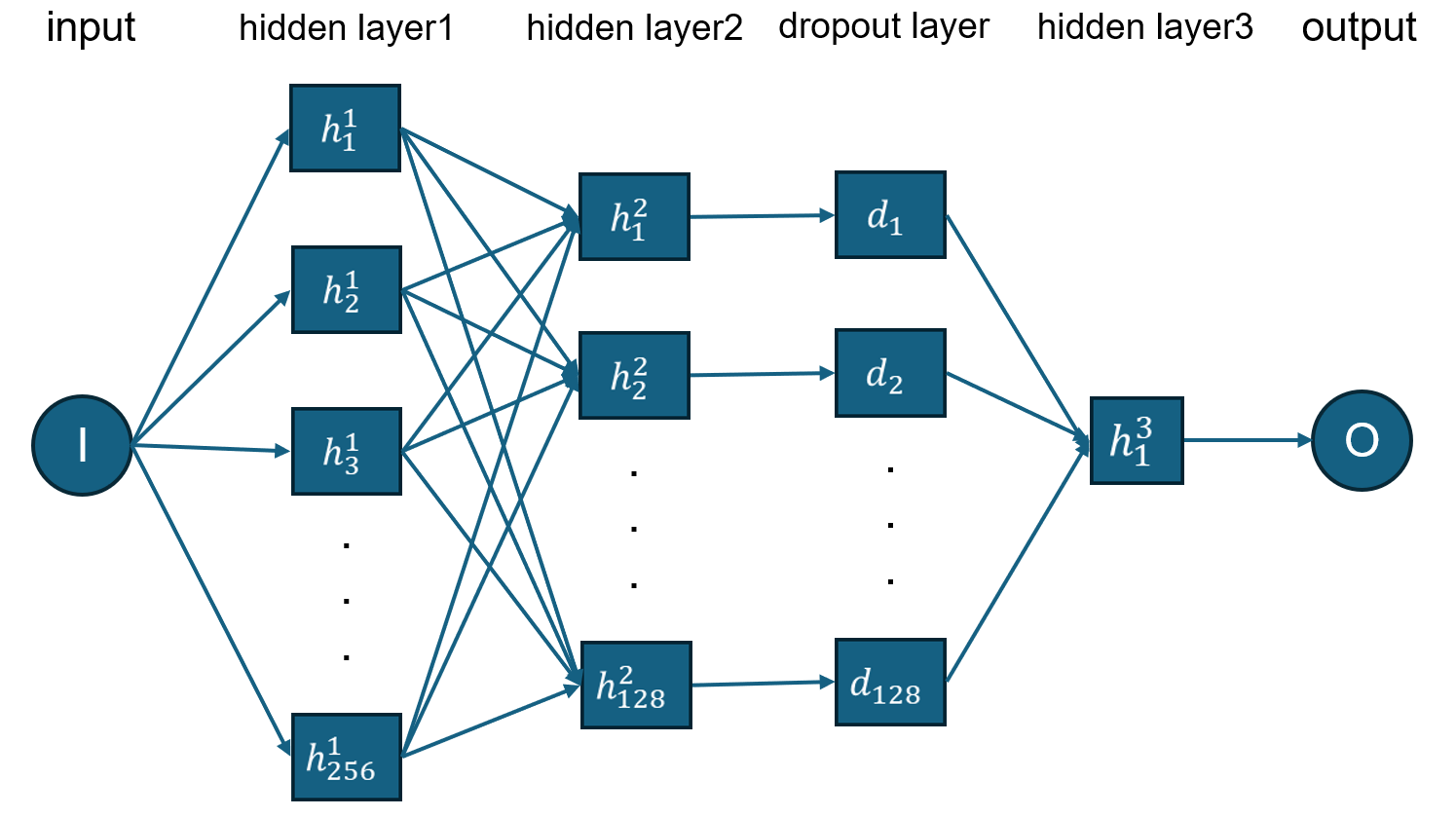}
    \endminipage
\caption{Model$_{K}$ architecture: estimation of $log\hat{K}$; $h^{j}_{n}$ represents $n^{th}$ neuron in $j_{th}$ hidden layer. This model takes 1 input feature (equation \ref{eq:9}). The dropout layer avoids overfitting of the model. The total number of parameters is 33K.}
\label{fig:k_reg_arch}
\end{figure}

The estimations of $\hat{\alpha}$ and $\hat{K}$ are performed with a ConvLSTM network (Model$_{\alpha}$, Figure~\ref{fig:alpha_reg_arch}) and a fully connected neural network (Model$_{K}$, Figure~\ref{fig:k_reg_arch}) respectively. The reason we estimate $\hat{\alpha}$ and $\hat{K}$ using two regression models in two different deep architectures rather than regress them together is that $\alpha$ and $K$ are independent terms. Even if we could estimate $\hat{\alpha}$ and $\hat{K}$ in a single architecture, it would make  the hyperplane unnecessarily much more complex than needed, since the number of parameters in the model would be higher. For the training, we generate 1~million observations without noise, 2-dimensional fBm sequences with uniformly selected lengths from 5 to 256. The model architectures are shown on Figures \ref{fig:alpha_reg_arch} and \ref{fig:k_reg_arch}. The models are extendable to higher-dimensional sequences for the estimation of both $\hat{\alpha}$ and $\hat{K}$. To optimise the models, we use RMSprop for Model$_{\alpha}$ and Adam~\cite{adam} for Model$_{K}$.


Since the convolutional layer accepts fixed length of input only, the trajectory sequence of length $T$ is divided into $N_{sub}$ sub-sequences where $N_{sub}$=$T$ mod $2^m + 1$ , $m$ is the maximum number which makes $2^m$ smaller or equal to $T$. We trained the models with $\mathbb{T}=\{5, 8, 12, 16, 32, 64, 128\}$. During the inference, the regressed values of $\alpha$ are averaged if $T-N_{sub}+1 \not\in \mathbb{T}$. The input features for the $\alpha$ regression are as follows:
\begin{equation}\label{eq:6}
	\displaystyle \mathbf{\alpha_1} = \bigg\{\frac{1}{\sigma_{\mathbf{X}}T}\sum_{t=0}^{j} \lvert x^{t+1}  - x^{t}\rvert\ \mid \ j\in [\, 0,\  T-1]\,  \bigg \} 
\end{equation}

\begin{equation}\label{eq:7}
	\displaystyle \mathbf{\alpha_2} = \bigg\{ \mathbf{R}^{j} / \overline{\mathbf{XY}}/{T}  \mid \ j\in [\, 0,\  T]\,  \bigg \}
\end{equation}

\begin{equation}\label{eq:8}
	\displaystyle \mathbf{\alpha_3} = \bigg\{ (\mathbf{X}  - \mathbf{X}^{0})/\overline{\mathbf{XY}}/{T} \bigg \}, 
\end{equation}
where $\mathbf{\sigma_{X}}$, $T$, $\mathbf{R}^k$ and $x^{t}$ represent the standard deviation of x coordinate sequence, length of input sequence, radial displacement for a sequence length at $k$ and the $t^{th}$ element in $\mathbf{X}$ respectively. Note that the first value of $\mathbf{\alpha_1}$ is padded with 0 for the equal length between features. These input features are for the estimation of $\hat{\alpha}$ in terms of $X$ space and we took a mean of $\hat{\alpha}$ for 2D trajectory computed for each dimension.

The input feature for $K$ regression is provided as:
\begin{equation}\label{eq:9}
	\displaystyle \mathbf{K_1} =\log\overline{\mathbf{XY}},
\end{equation}
where $\overline{\mathbf{XY}} = {\frac{1}{T}\sum\limits_{i=0}^{T-1} \sqrt{(x^{t+1} - x^{t})^2 + (y^{t+1} - y^{t})^2}}$ \ \ \ $k, T \in \mathbb{N}$.

\begin{figure}[!htb]
    \minipage{0.99\textwidth}%
    \centering
    \includegraphics[scale=0.65]{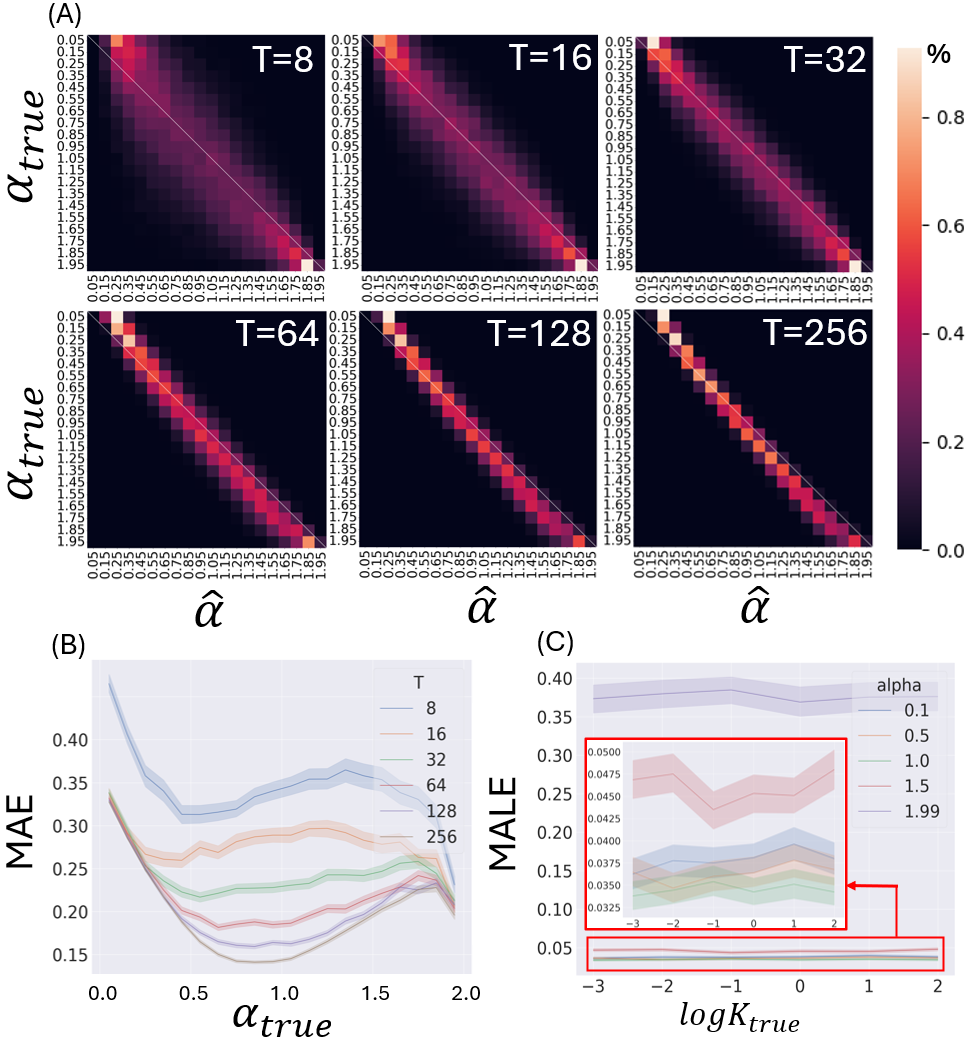}
    \endminipage
\caption{
(A): Regression results of $\hat{\alpha}$ on different sequence lengths $T$ with $\alpha_{true}\in\{0.05, 0.15, ..., 1.95\}$; $\hat{\alpha}$ is binned with window $0.1$ (to fit into the confusion matrix); 1,000 trajectories are simulated for each $\alpha_{true}$. (B): MAE (Mean Absolute Error) for $\alpha_{true}$ with different lengths $T$. The blurred region shows 95\% confidence interval. (C): $log\hat{K}$, MALE (Mean Absolute Logarithmic Error) for $logK_{true}$ with $\alpha_{true}\in\{0.1, 0.5, 1.0, 1.5, 1.99\}$. The red box contains the zoomed MALE values of $\alpha_{true}\in\{0.1, 0.5, 1.0, 1.5\}$. The blurred region shows 95\% confidence interval.
The estimation of $log\hat{K}$ becomes challenging, if $\alpha_{true}\in\{\alpha|\alpha>1.9\}$ due to the auto-covariance of fBm.
}
\label{fig:reg_result}
\end{figure}

For a stable estimation of $\hat{\alpha}$, we consider that $\mathbf{\alpha_1}$ is the most distinguishable input feature compared to others that represents the normalized absolute cumulative sum of fractional Gaussian noise. The numerical results of Model$_{\alpha}$ and Model$_{K}$ are shown in Figure \ref{fig:reg_result}. To test the models, we simulated 1,000 trajectories for each $\alpha_{true}$ and $K_{true}$. We can observe the the performance  decreases, if T gets smaller; in general this is due to the low number of observations and it leads possibly to a biased result, since fBm is a continuous Gaussian process. In the interval $\alpha_{true}=0.05$ to $\alpha_{true}=1.0$, $\sigma_{\hat{\alpha}}$ increases, and drops again from $\alpha_{true}=1.0$ to $\alpha_{true}=1.95$. For the longer trajectories, where $T>32$, we can see that the estimated results are the most promising in the neighbourhood near $\alpha_{true}=1.0$. The estimation of $\hat{K}$ shows reasonable results in general, however, if $\alpha_{true}$ gets bigger, such as $\alpha_{true} > 1.5$, the MALE criterion (equation \ref{eq:MALE}) increases rapidly due to the auto-covariance of fBm. In this case, accurate estimation of $\hat{K}$ is nearly impossible unless the first coordinates of a short trajectory are sampled near the true mean of the Gaussian distribution.

\subsection{Clustering of $\hat{\alpha}$ and $\hat{K}$ and sequential inference on a trajectory}

\begin{figure}[!htb]
    \minipage{0.99\textwidth}%
    \centering
    \includegraphics[scale=0.55]{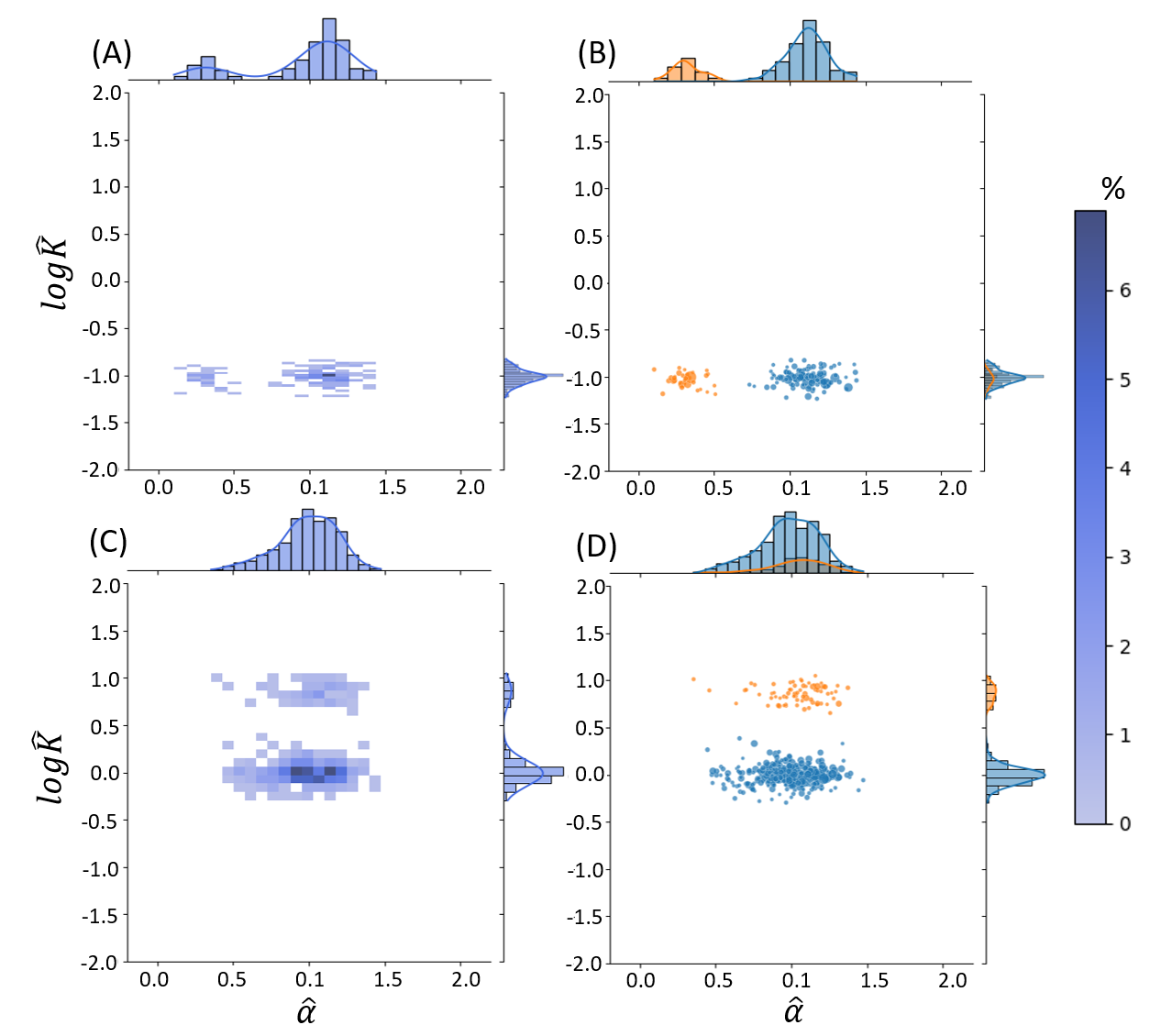}
    \endminipage
\caption{
Distributions of $\hat{\alpha}$ and $log\hat{K}$ of sub-trajectories regressed from a given sample. The scatter size corresponds to the normalized length of sub-trajectory. The colorbar represents the percent. (B), (D) show clustered $\hat{\alpha}$ and $log\hat{K}$ of the distribution (A) and (C) with GMM. (A), (C) are the 2-dimensional histogram with marginal distributions of $\hat{\alpha}$ and $log\hat{K}$. (A) and (B) are obtained from the sample where $\alpha_{true}$ changes between 0.3 and 1.0 while $K_{true}$ is 0.1. (C) and (D) are are obtained from the sample where $K_{true}$ changes between 0.8 and 1.0 while $\alpha_{true}$ is 1. Both samples are simulated with 1,000 trajectories where the transition probability is 0.01. The clustering with GMM is performed for the sub-trajectories where the length is longer than 16.
}
\label{fig:distribution}
\end{figure}

To filter the regression noise and false positive changepoints, we cluster $\hat{\alpha}$ and $log\hat{K}$ of sub-trajectories from a given sample. The clustering is performed with the Gaussian Mixture Models (GMM) for the sub-trajectories where the length $T_{sub}$ is longer than 16 due to high uncertainty of low number of observations. If the sample size is smaller, the length $T_{sub}$ of sub-trajectories for the clustering should be smaller to obtain enough data for the clustering. However, if the sample size is big enough, we can increase $\lambda$ to exclude uncertain short sub-trajectories. The optimal number of clusters $k$ is inferred with the BIC (Bayesian Information Criterion) score~\cite{BIC} or can be decided manually. BI-ADD produces the distribution of $\hat{\alpha}$ and $log\hat{K}$ shown in Figure \ref{fig:distribution} and the number of clusters can be deduced from the results.

With the Gaussian mixtures built in the pre-processing step, i.e. GMM clustering with estimated $\hat{\alpha}$, $\hat{K}$ of sub-trajectories on $\alpha$-$K$ space, we can determine that the approximated potential changepoints from a given trajectory are true positives or false positives. BI-ADD starts the inferences from the lowest value of changepoint in the transformed signal $\mathbf{S}$. If two consecutive sub-trajectories divided by the changepoint are estimated to belong to the same cluster with a given significance level (Table~\ref{table:params}) for each length of sub-trajectory, BI-ADD merges the selected two consecutive sub-trajectories and re-estimates $\alpha$ and $K$. The selection of clusters is determined by comparing the likelihoods of clusters constructed with GMM. BI-ADD repeats the inference and merges the sub-trajectories until BI-ADD does not detect any possible merging. The significance levels in Table~\ref{table:params} were fixed by a heuristic approach. These alternating iterative steps are performed sequentially.

\begin{table}
\caption{Significance level for each $T_{sub}$\label{table:params}}
  \begin{tabular}{l l l l l l l l}
     \hline
     sub-trajectory length($T_{sub}$)&  5&  8&  12&  16&  32&  64&  128\\
     \hline
     significance level& $10^{-5}$&  $10^{-3}$&  $2.5\times10^{-2}$&  $10^{-1}$&  $10^{-1}$&  $10^{-1}$&  $10^{-1}$\\
  \end{tabular}
  \end{table}

\section{Results}

To extensively test the proposed approach BI-ADD, we participated in AnDi2 Challenge as team SU-FIONA for both video task and trajectory task at individual trajectory-level. The simulated scenarios of AnDi2 Challenge aim to generate the molecular trajectories on micro or nanoscale by mimicking real world problems in biology. The details of comparative results between AnDi2 Challenge participants is available in \cite{andi_challenge_result}. For the video task, we applied  FreeTrace~\cite{FreeTrace} to predict the individual molecular trajectories from the video and analyzed the predicted trajectories with BI-ADD. For the inferred trajectories, BI-ADD was used to predict the changepoints and the diffusive properties of molecular trajectories. We were ranked 1st for the video task, and 6th for the trajectory task. 

\paragraph{Evaluation metrics}
We evaluate the performance of the method using the following metrics, also used by the AnDi2 challenge:

Mean absolute error between an estimation $\hat{y}_{i}$ and true value $y_{i}$:
\begin{equation}\label{eq:MAE}
	\displaystyle \textbf{MAE} = \frac{1}{N}\sum_{i} \left| \hat{y}_{i} - y_{i} \right|
\end{equation}

Mean absolute log error:
\begin{equation}\label{eq:MALE}
	\displaystyle \textbf{MALE} = \frac{1}{N}\sum_{i} \left| \log{\hat{y}_{i}} - \log{y_{i}} \right|
\end{equation}

Root mean squared error:
\begin{equation}\label{eq:RMSE}
	\displaystyle \textbf{RMSE} = \sqrt{\frac{1}{N}\sum_{i} \left( \hat{y}_{i} - y_{i} \right)^2}
\end{equation}

Jaccard Similarity Coefficient which measures the true positive ratio of changepoints along trajectory without considering true negatives:
\begin{equation}\label{eq:JSC}
	\displaystyle \textbf{JSC} = \frac{TP}{TP + FP + FN},
\end{equation}
where $TP$, $FP$ and $FN$ are the numbers of true positives, false positives and false negatives respectively.

Mean squared logarithmic error:
\begin{equation}\label{eq:MSLE}
	\displaystyle \textbf{MSLE} = \frac{1}{N}\sum_{i} \left( \log{(1 + \hat{y}_{i})} - \log{(1 + y_{i})} \right)^2
\end{equation}

\subsection{A two-state scenario of molecular trajectories: evaluation of $\alpha$ and $K$}

\paragraph{Data generation} We generate 20 000 2-dimensional fBm trajectories with T = 200. The two states of sub-trajectories are described by $\{(\alpha_1, K_1)$, $(\alpha_2, K_2)\}$, where the order of the states is random, and the transition probability between them is 0.02. 

\begin{figure}[!htb]
    \minipage{0.99\textwidth}%
    \centering
    \includegraphics[scale=0.48]{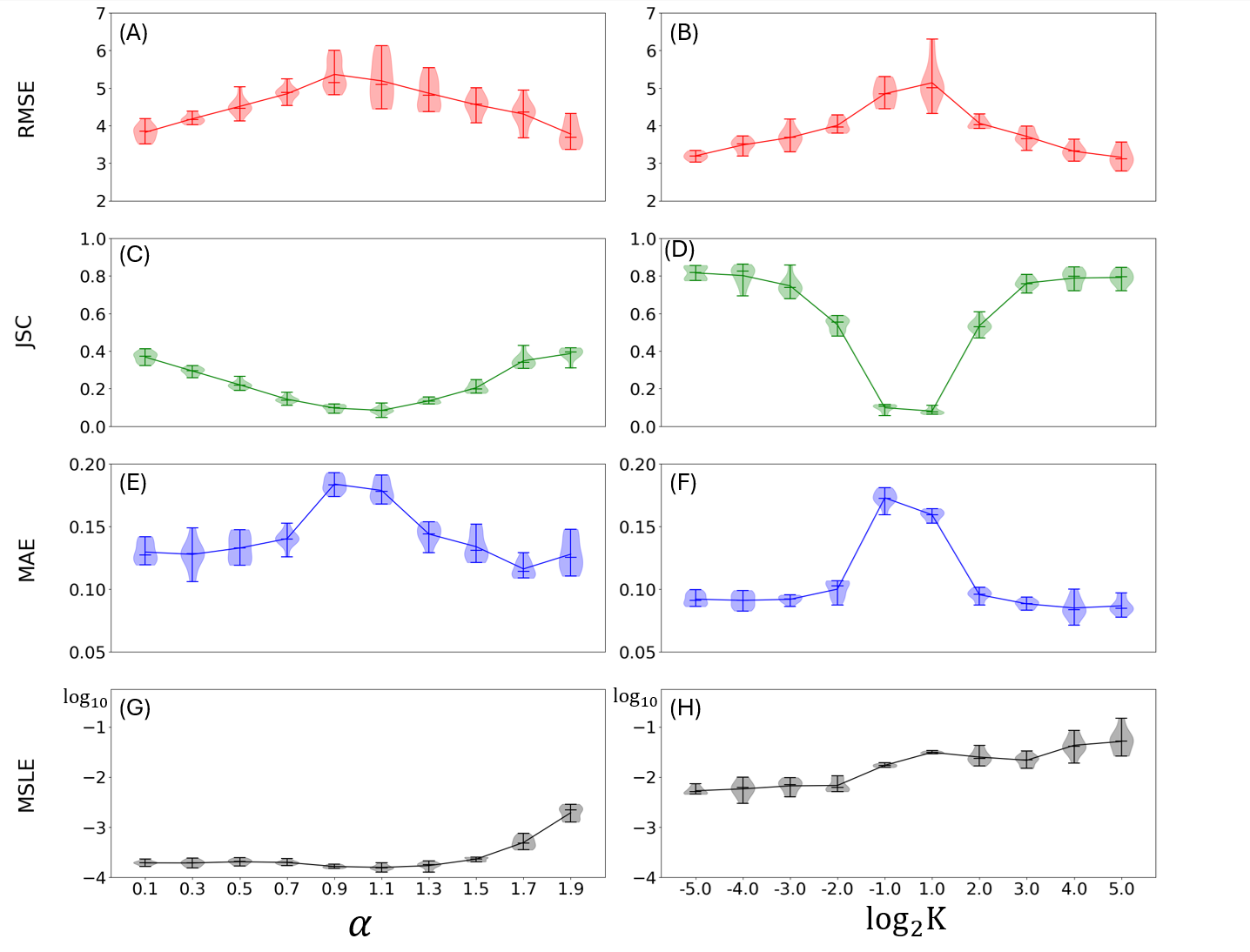}
    \endminipage
\caption{Performance of BI-ADD varying $\alpha$ and $K$ in different scenarios. On the left: 10 sets of scenarios $\{(0.1, \alpha), (0.1, 1.0)\ |\ \alpha\in\{0.1, 0.3, ..., 1.9\}\}$; on the right: $\{(log_2K, 1.0), (1.0, 1.0)\ |\ log_2K\in\{-5, -4, ..., 5\}\}$, where set=$\{(\alpha_1, K_1)$, $(\alpha_2, K_2)\}$. 1,000 fBm trajectories are simulated for each scenario with the transition probability 0.01. (A) and (B) show the RMSE (Root Mean Squared Error, equation \ref{eq:RMSE}) of true positive changepoints. (C) and (D) show the JSC (equation \ref{eq:JSC}) of changepoints. (E) and (F) show the MAE of true positive changepoints. (G) and (H) show the MSLE (Mean Squared Log Error, equation \ref{eq:MSLE}) of true positive changepoints. We can observe that if the difference of $\alpha$ or $K$ becomes smaller between two set of states, the errors increase in general. The MSLE is the highest, if $\alpha$=1.9 due to auto-covariance of fBm and the result is coherent with (C) of the Figure \ref{fig:reg_result}. The metrics are calculated with AnDi datasets~\cite{andi2_dataset}.
}
\label{fig:metrics}
\end{figure}

The $\alpha$ and $K$ for each state are defined by $\{(\alpha_1, K_1)$, $(\alpha_2, K_2)\}$. To explore the effect of various  $\alpha$ values, we created the sets with different $\alpha$ $\{(0.1, \alpha), (0.1, 1.0)\ |\ \alpha\in\{0.1, 0.3, ..., 1.9\}\}$ and $\{(log_2K, 1.0), (1.0, 1.0)\ |\ log_2K\in\{-5, -4, ..., 5\}\}$ to test the effect of $K$. The performance is assessed using the metrics mentioned above, the results are shown on Figure \ref{fig:metrics}. 

To measure the detection rate of $cp$ (frame of a changepoint) in a given sequence, we use the Jaccard Similarity Coefficient (JSC). We consider $\hat{cp}$ (an estimated frame of a changepoint) as true positive if $|\hat{cp} - cp_{true}| < 10$, and false positive otherwise. The obtained results are intuitive: when the difference between two states of molecular trajectories becomes smaller, BI-ADD underperforms compared to the case, where the difference between two states is significant. The identification of changepoints is more difficult in terms of $\alpha$ than $K$, and the maximum value of JSC is around 0.4 in Figure \ref{fig:metrics}-(C), where the difference of $\alpha$ between two states is at the maximum. We can see, however,  the maximum JSC is around 0.8 on Figure \ref{fig:metrics}-(D) which shows the detection of changepoints is easier in terms of $K$ than $\alpha$. The estimated results of $K$ on Figure \ref{fig:metrics}-(G) emphasize the difficulty of accurate $K$ estimation, especially, where $\alpha \simeq 1.9$ due to auto-covariance of fBm.

\subsection{A two-state scenario of molecular trajectories: the role of $T$}
\begin{figure}[!htb]
    \minipage{0.99\textwidth}%
    \centering
    \includegraphics[scale=0.48]{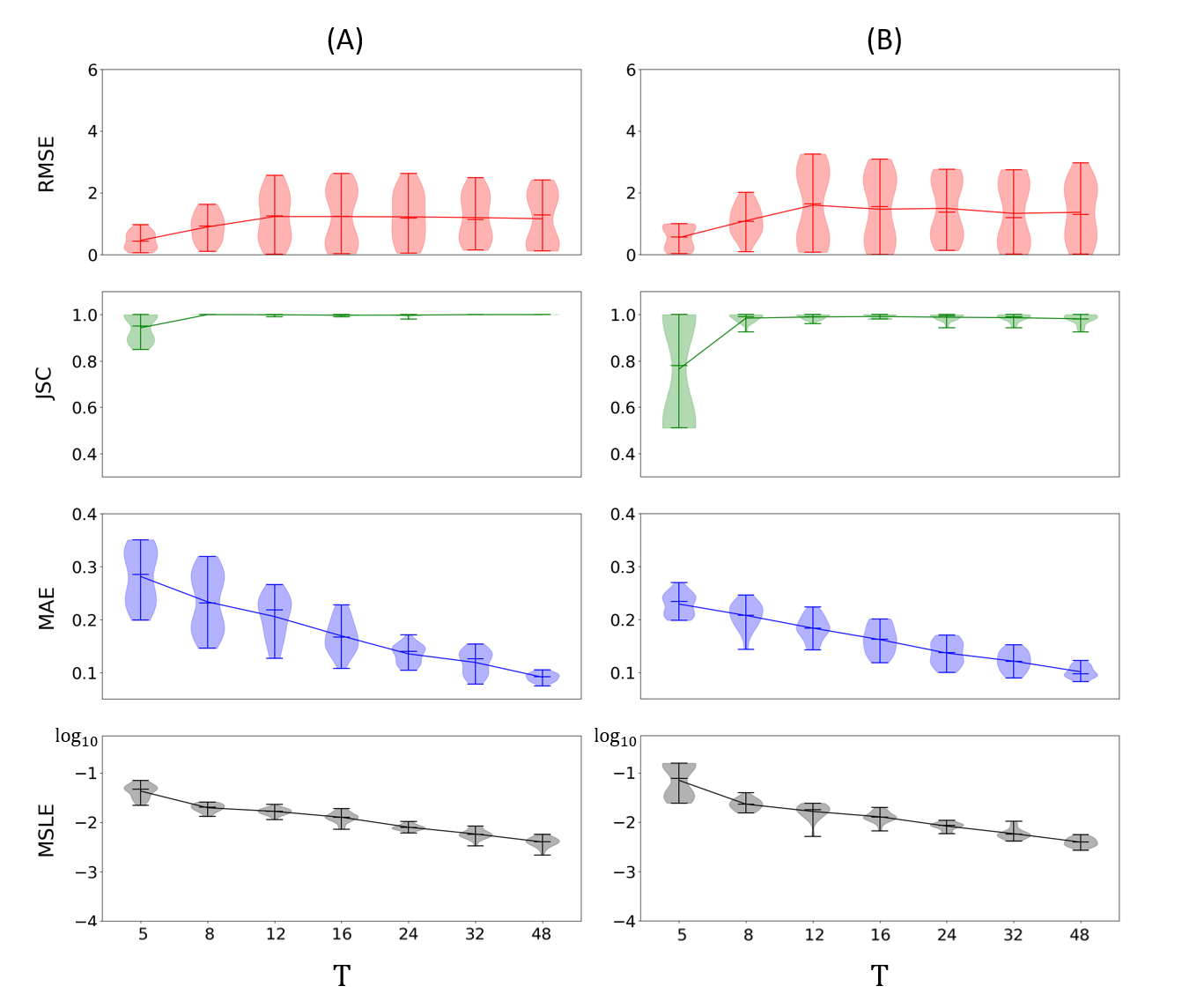}
    \endminipage
\caption{Performance of BI-ADD varying $T$ in two different scenarios. The simulated sets of ($K$,$\alpha$) of (A) are $\{(1.0, 1.0), (0.01, 0.2)\}$ and $\{(1.0, 1.0), (0.1, 0.5)\}$ for (B). Note that the simulated diffusivity-difference of (B) between two populations is smaller than (A). 500 trajectories of total length $2T$ are simulated for each $T$ with 1 changepoint in the middle of trajectories and 500 trajectories are simulated for each $T$ without changepoints to reflect the false negative sensitivity of BI-ADD. We can observe overall performance increases, if $T$ gets longer except RMSE, since the metrics are calculated only for true positives. The corresponding generated clusters for each $T$ is available in Figure S1. 
}
\label{fig:performance_over_time}
\end{figure}

The evaluation of BI-ADD for different length of short trajectories is shown in the Figure~\ref{fig:performance_over_time}. Scenario (A) is simulated with $\{(1.0, 1.0), (0.01, 0.2)\}$ and $\{(1.0, 1.0), (0.1, 0.5)\}$ for (B). BI-ADD performs well for both scenarios in general except for short trajectories where $T$ = 5. If the differences of clustered $\hat{\alpha}$ and $\hat{K}$ between two population are sufficiently large to make distinguishable clusters (Figure S1), the score of JSC for short trajectories such as $T$=5 is acceptable to use. If the distance between the generated clusters is not explicitly distinguishable, the variance of JSC is large for short trajectories and the identified changepoints might contain many false negatives and false positives.

\section{Conclusion and Discussion}
We introduced a novel method -- Bottom-up Iterative Anomalous Diffusion Detector (BI-ADD) -- to approximate the changepoints and estimate the diffusive properties $\alpha$ and $K$ from molecular trajectory. 
 The method is based on three central parameters of BI-ADD. The threshold ($\lambda$) of the transformed signal is set to 0.15 by default, $\lambda$ acts as a false negative controller, its low value can decrease the number of false negatives, but can increase the computational time, if it creates more sub-trajectories.  The automated estimation of $\lambda$ was not tackled in the current contribution, since it is a challenging topic which is out of score in this paper. However, this problem is among our future research directions.
 
 The size of the sliding windows ($w_i$) are set to $i\in\{20, 22, ..., 40\}$ by default. An optimal size of a sliding window is related to the transition probability between the states of trajectories: if the transition probability is low, the sliding window should not be small, since small windows are sensitive to the noise and the number of false positives will increase. The total number $k$ of the hidden states of a given sample is set to $-1$ by default, and an optimal $k$ is estimated with the BIC. However, the BIC criterion might produce a large number of clusters, if the distribution in Figure (\ref{fig:distribution}) contains many noisy sub-trajectories. This leads to multiple Gaussians and unnecessarily increases the number of estimated clusters $k$. Thus, manually choosing the number of clusters in real applications might be reasonable. 
 
 The memory storage and computational resources of BI-ADD are mostly consumed by the estimation of $\hat{\alpha}$ due to the size of $\text{Model}_{\alpha}$. The approximate computation time of BI-ADD mainly depends on the length of trajectories. The elapsed time for the scenario (A) in Figure~\ref{fig:performance_over_time} is available in Figure S3, the overall processing time of 1,000 trajectories is 522.9 seconds on a single machine equipped with RTX 3090, 24GB of VRAM. We trained the Model$_{\alpha}$ and Model$_{K}$ on a A100 chip with 1M simulated noiseless fBm 2-dimensional trajectories. BI-ADD infers the changepoint detection sequentially due to the merging mechanism. The sequential nature of the approach hinders parallelization of the changepoints approximation. A distributed detection of the changepoints for the molecular trajectories is an important research avenue. 

Another research direction is to tackle challenging scenarios which arise in real data. The BI-ADD copes extremely well with the analysis of molecular trajectory in cases, where the diffusive properties between the states are significantly different. If the difference of diffusive properties of a trajectory is small in terms of both $K$ and $\alpha$, BI-ADD performs sub-optimally, since this setting is particularly difficult for all methods. More precisely, the distributions of $\hat{\alpha}$ and $\log\hat{K}$ (e.g., Figure \ref{fig:distribution}) can be easily clustered since the clusters are well-separated in our numerical experiments. Note that the clustering is done using the Gaussian Mixture Models. In our contribution, we make use of the obtained clustering, however, we did not explore yet the transition probabilities between clusters. If the estimated transition probabilities between the states are time dependent, this change over time can be studied for the biological inter-molecular processes. Another open problem is an accurate estimation of $\hat{\alpha}$ from a short trajectory, where the length is smaller than 8. In a setting where a lot of sequences are short, it becomes hard to perform the estimation accurately due to the low number of observations. Since the accurate estimation of $\hat{\alpha}$ and $\hat{K}$ is nearly intractable for very short trajectories, the application BI-ADD for short trajectories in real data should be decided after analysis of the obtained clusters (such as the clusters in Figure S1). This point is critical for the prediction of molecular trajectories in biology, since inferring long trajectories from real data is limited in empirical methods of optical physics and the data acquisition is, in general, expensive.

We are currently testing the proposed BI-ADD on the experimental data of the FIONA team, Sorbonne university.

\section{Software availability}
The Python source code of BI-ADD is publicly available for research purposes.~\cite{BI-ADD}(\url{https://github.com/JunwooParkSaribu/BI_ADD})

\printbibliography 

\end{document}


\title{
\Huge{Supplementary information}\\
\Large{Bottom-up Iterative Anomalous Diffusion Detector (BI-ADD)}
}

\author[1\thanks{Corresponding author}]{Junwoo Park}
\author[1]{Nataliya Sokolovska}
\author[2]{Cl\'ement Cabriel}
\author[2]{Ignacio Izeddin}
\author[1]{Judith Min\'e-Hattab}
\affil[1]{\small{Sorbonne Universit{\'e}, CNRS, Laboratoire de Biologie Computationnelle et Quantitative, LCQB, F-75005 Paris, France}}
\affil[2]{\small{Institut Langevin, ESPCI Paris,
Universit{\'e} PSL, CNRS, Paris, France}}

\maketitle

\begin{figure}[!htb]
    \minipage{0.99\textwidth}%
    \centering
    \includegraphics[scale=0.48]{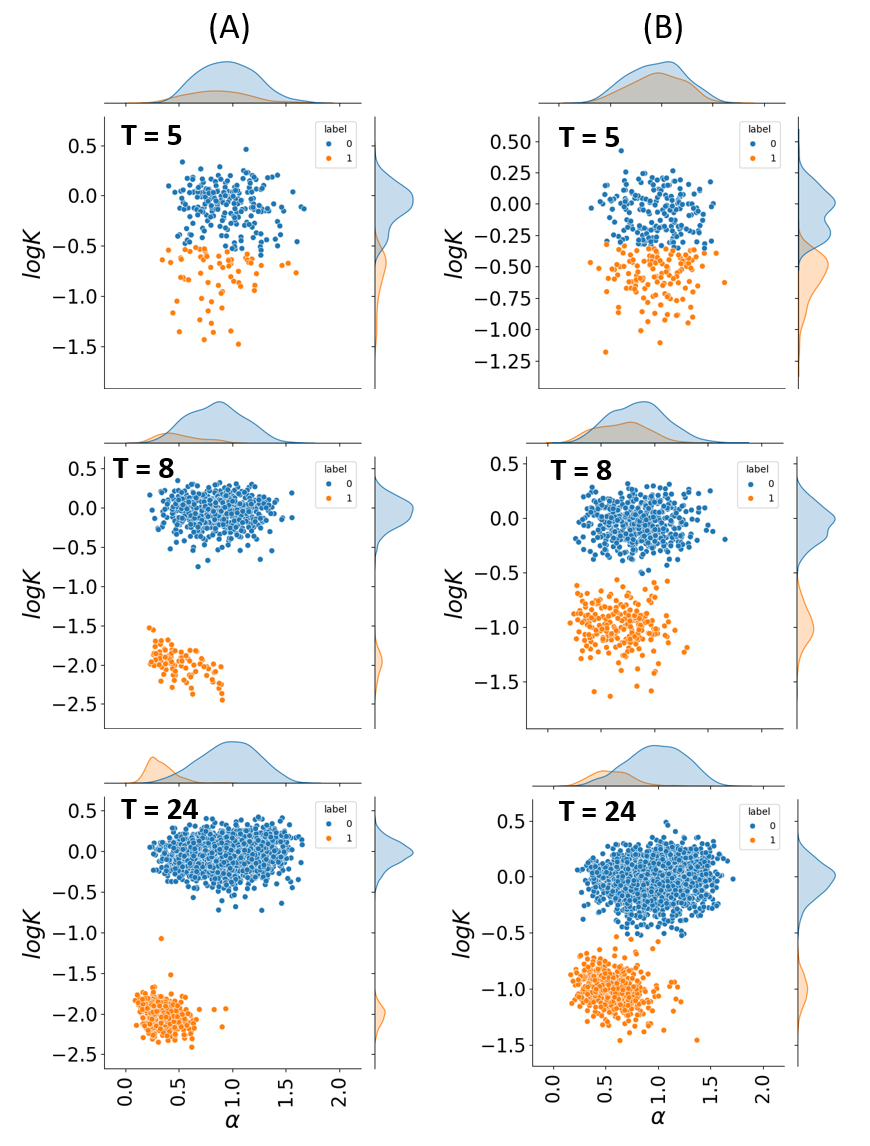}
    \endminipage
\caption{Generated clusters for $T$=5, 8 and 24 utilizing BI-ADD. ($\{(1.0, 1.0), (0.01, 0.2)\}$ for the column (A) and $\{(1.0, 1.0), (0.1, 0.5)\}$ for the column (B), which are the same scenarios of Figure 9. The distinction between two clusters is not clear for short trajectories($T$=5) in both scenarios. If the true diffusivity between two clusters are not distant(column B with $T$=5), the error of JSC is high compared to distant clusters(column A with $T$=5) for the same length of short trajectories. The means of clusters get closer to the true $\alpha$ and $K$ when T gets longer. The utilization of BI-ADD for short trajectories should be decided after observing the generated clusters.  
}
\label{fig:supp_1}
\end{figure}

\begin{figure}[!htb]
    \minipage{0.99\textwidth}%
    \centering
    \includegraphics[scale=0.55]{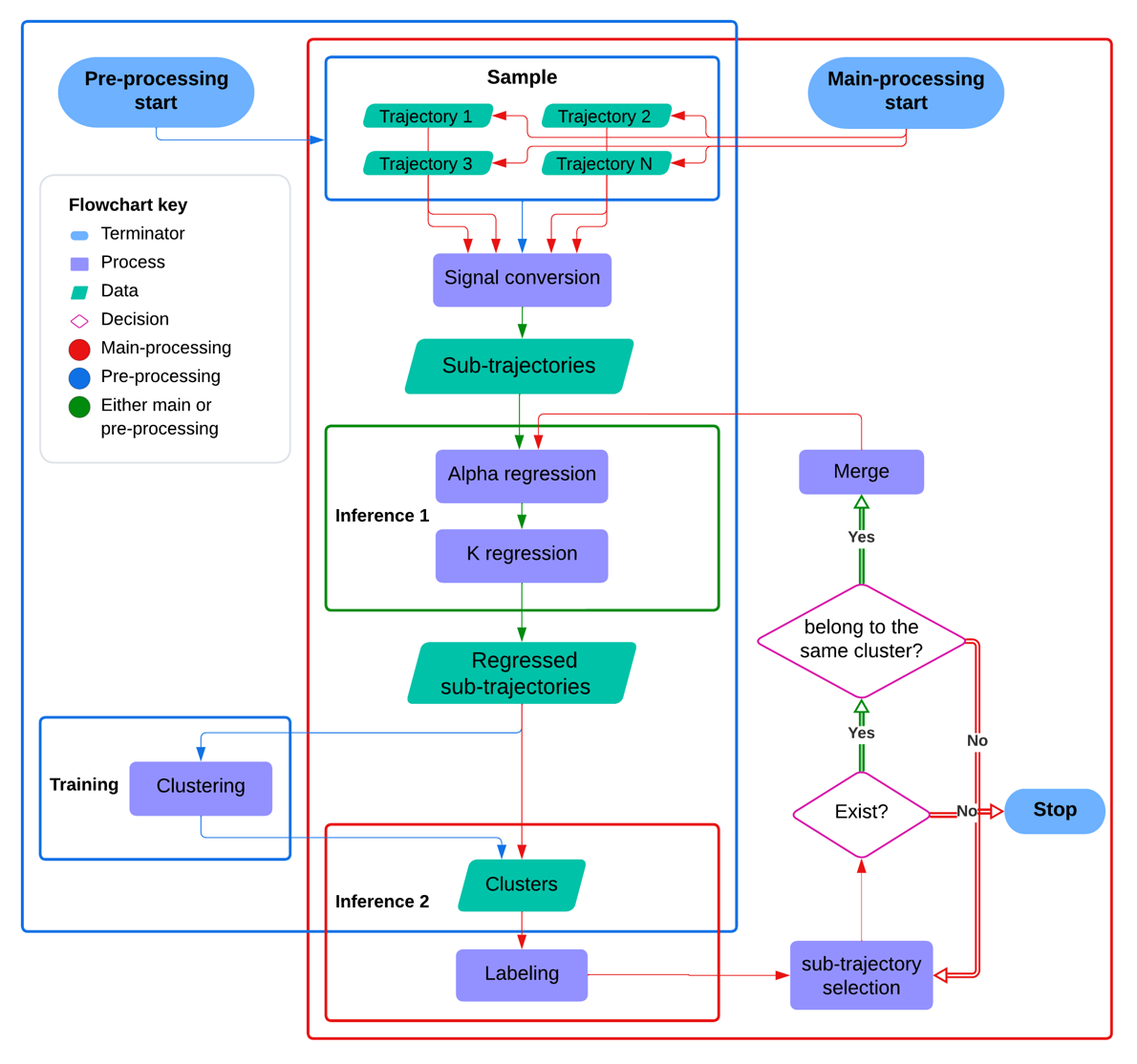}
    \endminipage
\caption{Detailed workflow of BI-ADD. In the pre-processing step, BI-ADD generates clusters from a given sample on $\hat{\alpha}$ and $log\hat{K}$ space. In the main processing step, the selection process contains an ordering of the potential changepoints in the order of lowest value from the normalized signal $\mathbf{S}$ converted with the equations (3) to (7), and selects two first non-selected consecutive sub-trajectories. If two selected consecutive sub-trajectories belong to the same cluster with a given significance level, BI-ADD merges the sub-trajectories and repeats the process.}
\label{fig:full_flowcharts}
\end{figure}

\begin{figure}[!htb]
    \minipage{0.99\textwidth}%
    \centering
    \includegraphics[scale=0.42]{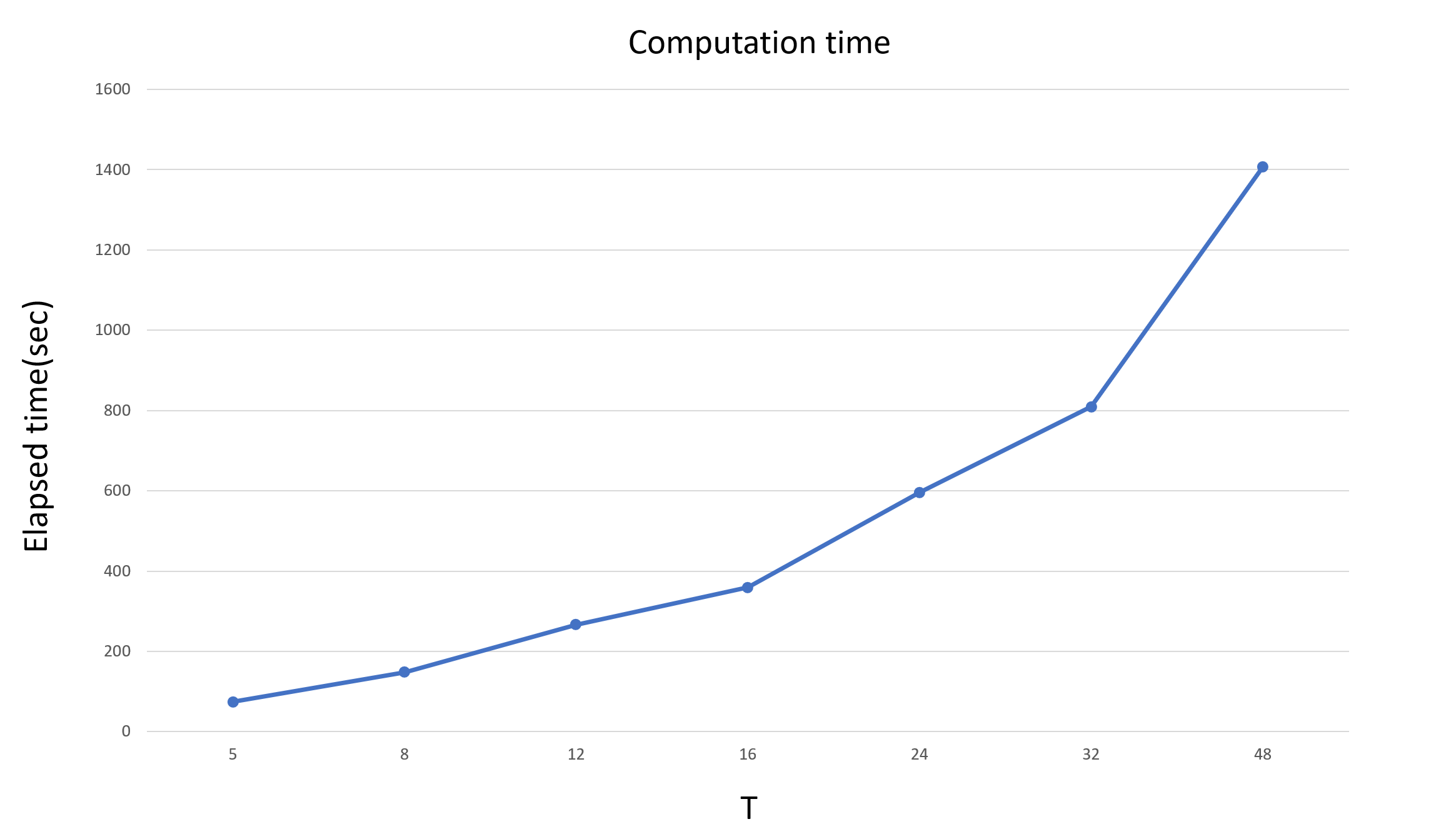}
    \endminipage
\caption{Computation time of BI-ADD for the scenario of Figure 9-(A). 1,000 trajectories are simulated for each $T$ and this computation time includes every process including pre- and main-processing. Inference of trajectories is performed with a single GPU RTX 3090, 24GB VRAM chip.}
\label{fig:full_flowcharts}
\end{figure}